\UseRawInputEncoding
\documentclass[10pt, conference]{IEEEtran}
\IEEEoverridecommandlockouts
\usepackage{float}
\usepackage{cite}
\usepackage{amsmath,amssymb,amsfonts}
\usepackage{algorithmic}
\usepackage{graphicx}
\usepackage{textcomp}
\usepackage{pgfplots}
\usepackage{booktabs}
\usepackage{subcaption}
\pgfplotsset{compat=1.17}

\def\BibTeX{{\rm B\kern-.05em{\sc i\kern-.025em b}\kern-.08em
    T\kern-.1667em\lower.7ex\hbox{E}\kern-.125emX}}
\begin{document}

\title{Assessing Privacy Preservation and Utility in Online Vision-Language Models}

\author{\IEEEauthorblockN{Karmesh Siddharam Chaudhari, Youxiang Zhu, Amy Feng, Xiaohui Liang, and Honggang Zhang}
		\IEEEauthorblockA{Department of Computer Science\\
	University of Massachusetts Boston, MA, USA
		}
		Email: \{K.Chaudhari004, Youxiang.Zhu001, Xiaohui.Liang, Honggang.Zhang\}@umb.edu, 
  amyjfeng1@gmail.com
	}

\maketitle

\begin{center}
\small
Accepted for publication in IEEE ICC 2026.\\
\copyright\ IEEE.\\
The final version will appear in IEEE Xplore.
\end{center}

\begin{abstract}



The increasing use of Online Vision Language Models (OVLMs) for processing images has introduced significant privacy risks, as individuals frequently upload images for various utilities, unaware of the potential for privacy violations. Images contain relationships that relate to Personally Identifiable Information (PII), where even seemingly harmless details can indirectly reveal sensitive information through surrounding clues. This paper explores the critical issue of PII disclosure in images uploaded to OVLMs and its implications for user privacy. We investigate how the extraction of contextual relationships from images can lead to direct (explicit) or indirect (implicit) exposure of PII, significantly compromising personal privacy. Furthermore, we propose methods to protect privacy while preserving the intended utility of the images in Vision Language Model (VLM)-based applications. Our evaluation demonstrates the efficacy of these techniques, highlighting the delicate balance between maintaining utility and protecting privacy in online image processing environments.  
\end{abstract}


\section{Introduction}
With the rapid advancement of Online Vision-Language Models (OVLMs), users increasingly upload images to these models for a wide range of tasks, including content generation, visual analysis, and decision support. Prior research has shown that the widespread deployment of vision-language models in real-world domains such as education, healthcare, and communication significantly increases the surface area for potential privacy violations \cite{li2024papillon, mireshghallah2024wildchat}. At the same time, the best-performing OVLMs are predominantly proprietary, cloud-hosted platforms accessible only through web interfaces or APIs (e.g., GPT-5, Claude 3.5, Gemini 2.5 Pro), which limits transparency and user control. This growing reliance on OVLMs introduces substantial privacy risks, as uploaded images may inadvertently disclose Personally Identifiable Information (PII). Prior studies also demonstrate that modern vision-language models can infer sensitive attributes such as occupation, location, or affiliations even from seemingly innocuous images \cite{staab2024privacy, wildvision2024}. Crucially, these privacy risks extend beyond explicitly identifiable elements such as faces or textual content. Sensitive information may also be inferred implicitly through contextual relationships among objects, locations, and background details within an image. Such \emph{implicit privacy leakage} underscores the complexity of privacy threats in image-based interactions with VLMs, as sensitive user information is often not directly observable but emerges from contextual visual cues \cite{zhu2024lppa, wildvision2024}.

 Therefore, simple objects within images, when considered in context, can reveal sensitive details that compromise privacy. This phenomenon has been further validated in image-to-text privacy analysis benchmarks like WildVision, which highlight the inference risks from image context \cite{wildvision2024}. In this paper, we propose and discuss strategies, such as the removal and masking of privacy-sensitive objects, to mitigate the privacy risks in an image. While effective to some extent, these approaches introduce a fundamental trade-off: reducing the potential for PII disclosure often comes at the cost of degrading the utility of the image for its intended purpose in VLM applications.

The challenge becomes even more pronounced in practical scenarios. Images are inherently complex, contain numerous objects with varied and sometimes subtle relationships to PII. Users often struggle to manually assess the privacy implications of each element or understand how modifications affect utility. This lack of granularity and guidance leaves users with limited options, making it difficult for them to make informed decisions about the privacy-utility trade-off.

In this paper, we propose a system that enables enhanced user control over image privacy while preserving utility. The system consists of three modules that jointly quantify privacy and utility, supporting informed trade-offs when applying removal and masking strategies.

First, we leverage VLMs to analyze and measure the relationship between an image and PII. This method identifies not only explicitly (directly) recognizable objects/information but also implicit (indirect) patterns that may reveal sensitive information indirectly. 

Second, we introduce a local model that infers the utility impact of VLMs by comparing original and modified images along with their descriptions, allowing users to estimate the usefulness of images after applying privacy modifications. Finally, we develop an intuitive user interface that visualizes privacy gains and utility impacts, allowing users to make informed decisions when interacting with online VLMs.

By integrating these modules, our system provides a practical and user-friendly approach to balancing privacy and utility in image-based VLM applications, addressing critical gaps in existing solutions. 
\vspace{-0.9em}
\begin{figure*}[h]

\centering
\includegraphics[width=0.90\textwidth, keepaspectratio=false]{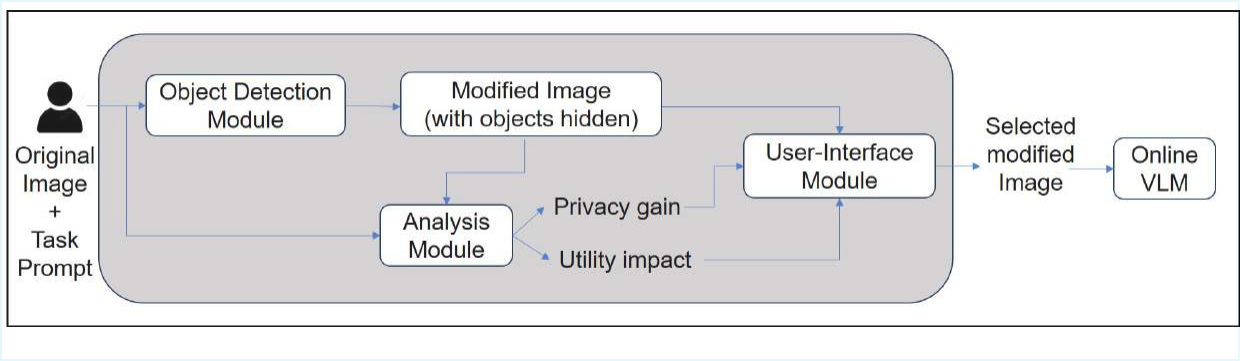}
  \caption{PPA system}\label{fig:system_model}
\end{figure*}

\section{Related work}
Recent advancements in OVLMs have expanded their capabilities in understanding images, but they have also amplified concerns regarding privacy. As users upload images for various tasks, it reveals PIIs, both explicit and implicit.
Privacy inference research by Tömekçe et al. \cite{staab2024privacy} provides compelling evidence that vision-language models can infer sensitive attributes such as location, occupation, and interests from seemingly innocuous images, even in the absence of human subjects. Their study reports inference accuracies as high as 77.6\%, and demonstrates that standard VLM safety filters can be bypassed through prompt engineering and image manipulation techniques. Similarly, Lu et al.'s WildVision benchmark \cite{wildvision2024} reveals how VLMs are prone to hallucinations and privacy leakage in real-world image settings.

These findings underscore the growing concern that modern VLMs—despite their advanced capabilities—can expose sensitive information not just from training data, but also through inference on user-generated inputs \cite{smith2024privacyllm, xue2023dpimage, yu2021ganprivacy, mirjalili2020privacynet}. Prior approaches have explored structured privacy control through differential privacy in image feature space \cite{xue2023dpimage}, GAN-based visual de-identification \cite{yu2021ganprivacy}, and semi-adversarial methods for attribute masking \cite{mirjalili2020privacynet}. This motivates the need for our proposed system, which addresses both explicit and implicit privacy risks by masking or removing potentially sensitive objects prior to uploading images to OVLMs.

LLMs also present similar challenges: prior work by Carlini et al. \cite{carlini2021extracting, carlini2023extracting} has shown that these models can memorize and regenerate verbatim training data. Moreover, Mireshghallah et al. \cite{mireshghallah2024wildchat} highlight that even casual prompts to ChatGPT can lead to unintended PII disclosures, further emphasizing the importance of privacy-preserving mechanisms.
Zhu et al. proposed a Local Privacy-Preserving Prompt Assistant (LPPA), showing that prompt rewriting and masking could significantly reduce leakage with minimal loss in utility \cite{zhu2024lppa}. Their approach is closely aligned with our obfuscation-based image modification technique.
Li et al. proposed PAPILLON, a hybrid delegation system combining local and remote LLMs for privacy-preserving reasoning. It highlights the trade-offs in privacy and utility when the partial delegation is required for task completion \cite{li2024papillon}.

More advanced approaches include DP-Image, which applies differential privacy to image feature vectors \cite{xue2023dpimage}, and PrivacyNet, which allows selective suppression of soft-biometric attributes in face images while preserving recognition utility \cite{mirjalili2020privacynet}. These go beyond naïve obfuscation and enable structured privacy control.

The amplifying reliance on online VLMs for image and data processing has increased significant privacy concerns. Previous research and work has explored various aspects of privacy risks in digital images, focusing on PII exposure, mitigation strategies, and the trade-off between privacy and utility. However, existing solutions do not emphasize implicit privacy risks and often struggle to address explicit privacy risks accurately.
\vspace{-0.5em}
\section{System model}    

We propose a Privacy Preserving Assistant (PPA), a local system that runs on a user's device. The user’s original prompt and uploaded image stay locally, while the modified images and prompts are sent to the OVLMs. The modified images do not contain selected objects with respect to the computed privacy score based on explicit or implicit sensitive information, thus preserving privacy. However, using the modified prompt may result in a different OVLM output. To help the user choose the modified image and the prompt the PPA generates a set of metrics associated with each prompt, including privacy techniques, privacy gain, and utility impact. As shown in Fig. 1, the PPA consists of an Object Detection Module (ODM), an Analysis Module, and an User Interface Module (UIM).

The \textbf{object detection module} aims to identify and generate multiple modified images based on user's input task prompt and input image denoted as $\ I_{u}$ and $\ I_{i}$ respectively, using different modification techniques.
 
\textbf{Explicit Privacy Risks}: These occur when sensitive information is directly accessible or extractable from the PIIs in the OVLM’s outputs. For each $'obj'$ identified, ODM assesses and quantifies the likelihood that it contains sensitive information which is directly extractable from the provided input image (e.g. PIIs like location) with a privacy risk score $P_{exp}{obj}$.

\textbf{Implicit Privacy Risks}: These occur when sensitive information is indirectly derived or extracted from the combination of existing/available PIIs in the OVLM output. Also, these can be the result of the explicit privacy leaks, i.e. a combination of the explicit privacy leaks.

The ODM module takes an input image,$\ I_{i}$, and an input task prompt,$I_{u}$. It detects all objects present in the image, denoted as $'obj'$. Let $'n'$ represent the total number of objects detected in $I_{i}$.
The object detection module is responsible for generating multiple modified images, denoted $Imod_{i}$. To achieve this, the system first uses an object recognition model to detect sensitive objects in $\ I_{i}$ based on a predefined set of privacy-sensitive categories  $\{Imod_{i}|1\leq i \leq n_{sen}\}$ (e.g. location, education, occupation) and the privacy module as $\ f_{p}$($\ I_{i}$) = $\{Imod_{i}|1\leq i \leq n_{sen}\}$. These categories are informed by prior privacy as such as those introduced in PUPA and WildChat and advanced object-relation modeling from scene graph–based reasoning approaches like G2(visual commonsense reasoning
generation method) \cite{li2024papillon, mireshghallah2024wildchat, yuan2025g2}.
The \textbf{analysis module} evaluates each modified image $\ Imod_{i}$ compared to the original image$\ I_{i}$ to generate a set of metrics $\{m_{i,j} | 1 \leq j \leq k\}$, where k represents the number of metrics used to analyze each prompt.



The \textbf{UIM} provides an intuitive visualization of the modified images $\{ Imod_{i} | 1 \leq i \leq n_{sen}\}$ and their associated metrics $\{m_{i,j} | 1 \leq j \leq k\}$. Users can rank the modified prompts according to the privacy  $G_{p}$ or the utility impact $U_{i}$. This allows users to easily compare and select a modified image that offers the best balance between privacy preservation and utility retention.

\section{PPA implementation}

This section presents a detailed implementation of the ODM, Analysis Module, and UIM.
\vspace{-0.2em}
\subsection{Object Detection Module}
 
In the object detection module, we propose two modification techniques to generate modified images from the user's original image, as shown in Fig. 2. First, we define the set of $\ C_{sen}$ pre-defined privacy-sensitive categories $\{C_{i}|1\leq i \leq C_{sen}\}$, including but not limited to PIIs(location, occupation, etc.) so that users can flexibly select and incorporate new sensitive categories relevant to their application of the system using the UIM. Then, based on these categories, we detect privacy-sensitive objects in the user's original image $\ I_{i}$  using an \textbf{object recognition model}. The detected objects, belonging to category $\ C_{i}$, are denoted as $\ S_{obj}$. Finally, we apply one of the following modification techniques to mitigate privacy risks while generating a modified image $Imod$:

    \begin{itemize}
        \item \textbf{Remove Technique}: A sensitive object $\ S_{obj}$ is completely removed from $\ I_{i}$. The modified image becomes:\[I_{mod}=\ I_i - S_{obj}.\] 

        \item \textbf{Mask Technique}: The sensitive object $S_{obj}$ is visually replaced with a placeholder representing its category $C_i$ within the image, resulting in the modified image 
\[
I_{mod} = I_i - S_{obj} + C_i.
\]
    \end{itemize}
In practice, object removal is implemented via region blurring to preserve visual coherence; accordingly, we treat blurring as a realizable form of object removal in our evaluation.
The object detection module outputs modified images, covering the various modifications across one or all detected sensitive objects. These are aligned with traditional methods but differ from GAN-based frameworks that seek to reconstruct images with de-identified features \cite{yu2021ganprivacy, xue2023dpimage}, which may provide higher utility in controlled settings. In this paper, we mainly focus on hiding sensitive objects to an extent where we could demonstrate the implicit privacy risk, which is not easily distinguished by humans but is measurable by a tool. 
\begin{figure}[h]
    \centering
    \centering\includegraphics[width=0.51\textwidth, keepaspectratio=false]{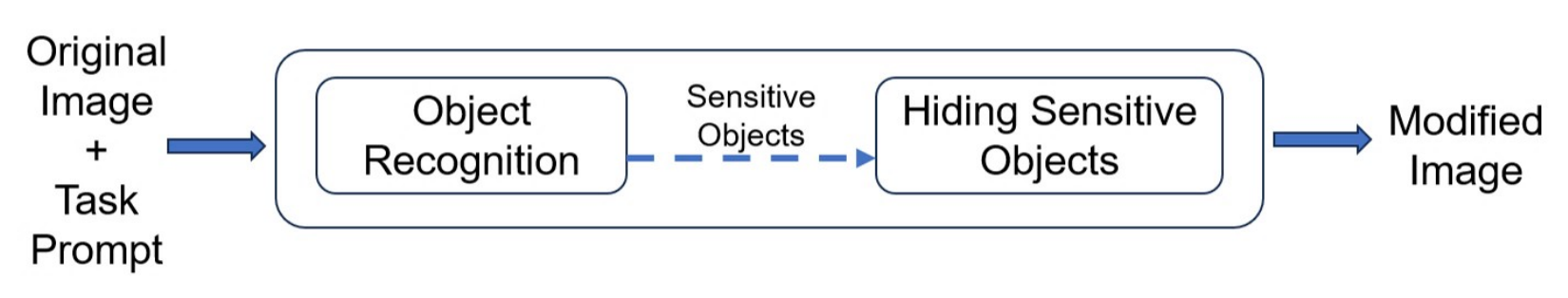}
    \caption{Object Detection Module}
    \label{fig:enter-label}
\end{figure}

\begin{enumerate}

\item {
By using the remove technique a sensitive object from the input image $I_{i}$ to reduce identifiable visual cues (e.g. location indicators). While this enhances privacy, the resulting image may cause the OVLM to generate incomplete or contextually inaccurate responses due to missing content.}
\item{
By using the mask technique a sensitive object in $I_{i}$ is replaced with a neutral placeholder from category $C_{i}$, preserving the image's structure but concealing specific details. This can lead the OVLM to produce vague or misinterpreted outputs, as masked features may introduce ambiguity into the visual context.}
\end{enumerate}

\vspace{-0.2em}
\subsection{Analysis Module}
The analysis module aims to analyze the impact of modifications by comparing the original and respective modified responses. The module performs the following analyses, as shown in Fig. 3:
\begin{figure}[h]
    \centering
        \centering\includegraphics[width=0.48\textwidth, keepaspectratio=false]{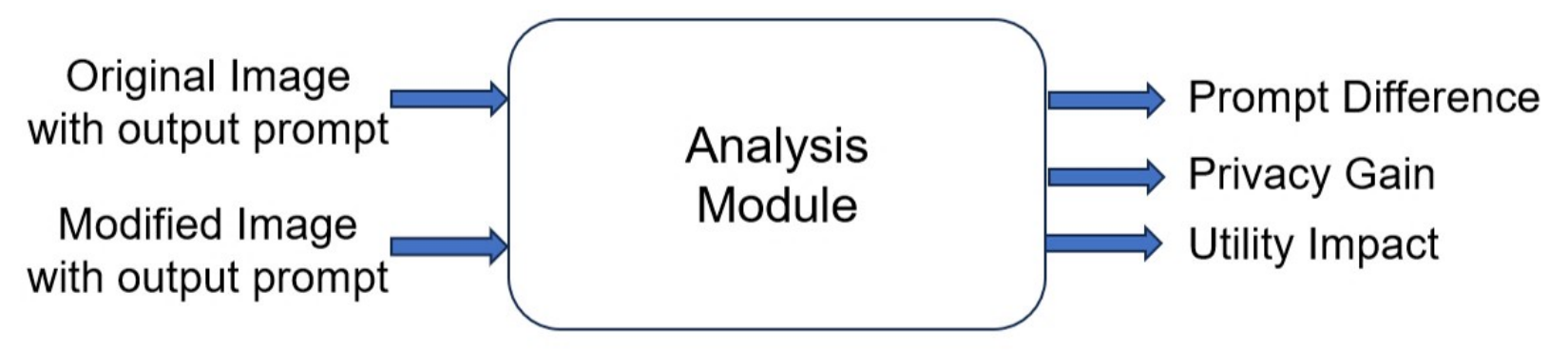}
         
    \caption{Analysis component}
    \label{fig:enter-label}
\end{figure}

\begin{enumerate}
        
    \item \textbf{Prompt Difference}: The module calculates the semantic/embedded similarity and the number of changes between the original prompt response denoted as $\ R_{orig}$ and each modified prompt response denoted as $\ R_{mod}$ with respect to the removal/masking of sensitive objects to show the extent of the modification. This metric echoes the prompt-level utility evaluations in LPPA \cite{zhu2024lppa} and delegation-based quality assessments in PAPILLON \cite{li2024papillon}.
    This is essential for users to understand the impact of the modification.
        
    \item \textbf{Privacy Gain:}
Privacy gain quantifies the reduction in sensitive information in a modified response relative to the original. It is defined as: \[
G_p(R_{\text{mod}}) = P(R_{\text{orig}}) - P(R_{\text{mod}})\] where \(P(R) \in [0,1]\) denotes the normalized privacy leakage score of response \(R\), computed based on the presence of PII-related content. \(R_{\text{orig}}\) is the response generated from the original image, and \(R_{\text{mod}}\) is the response after applying privacy-preserving modifications. A higher privacy gain indicates an effective privacy reduction.
        
\begin{figure*}[t]

\centering
\includegraphics[height=4cm,width=0.83\textwidth, keepaspectratio=false]{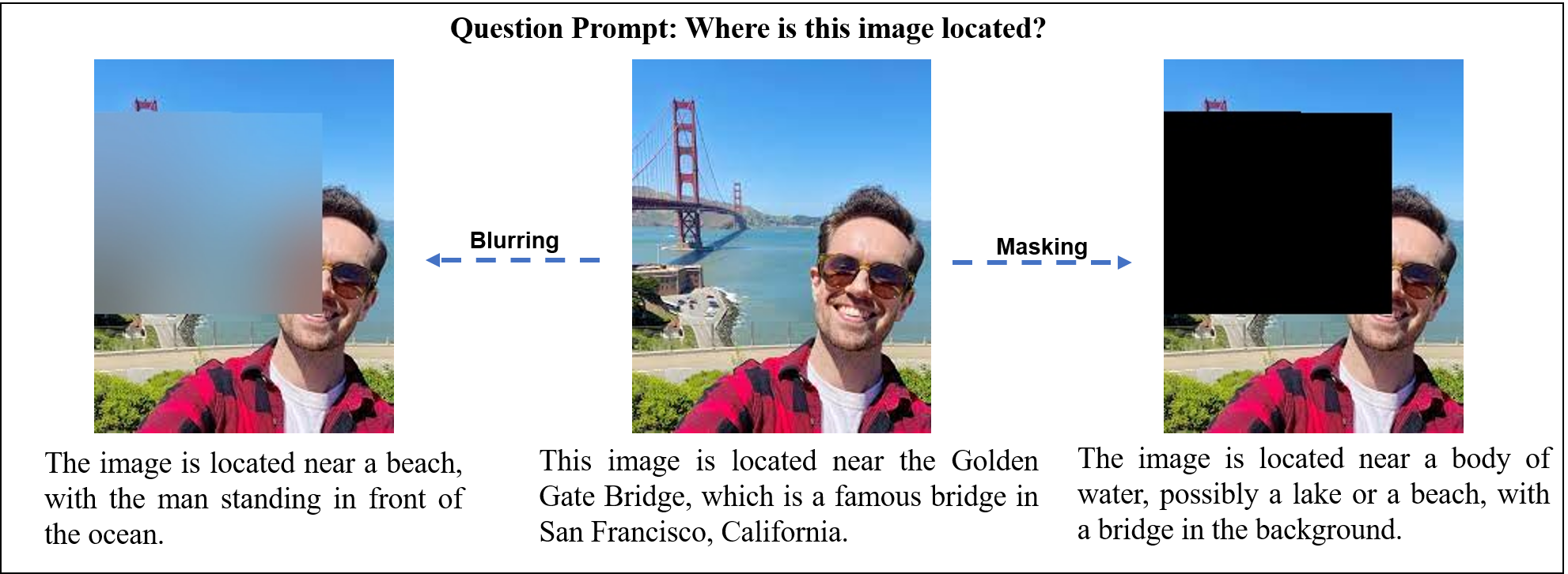}
  \caption{Responses to \textit{``Where is this image located?''} show how obfuscation alters location-specific PII inference.
}\label{fig:system_example}
  \vspace{-0.5cm}
\end{figure*}        

    \item\textbf{Utility Impact:}
Utility impact measures the loss in response usefulness due to modification and is defined as:\[U_i(R_{\text{mod}}) = 1 - U(R_{\text{mod}})\] where \(U(R_{\text{mod}})\in[0,1]\) is the embedding-based semantic similarity (cosine similarity between sentence embeddings) of \(R_{\text{orig}}\) and \(R_{\text{mod}}\). Since the original response serves as the reference, its baseline similarity is defined as 1. Thus, a higher \(U_i(R_{mod})\) indicates greater degradation of the original intent or response relevance.

\end{enumerate}
Privacy and utility are measured using distinct metrics. These two measurements are independent but jointly characterize the privacy–utility trade-off.

\vspace{-0.5em}
\subsection{User-Interface Module:}

The UIM presents key outputs from the analysis module to assist users in making informed privacy–utility decisions. It visualizes the trade-offs between privacy gain and utility impact for each modified output. The original image $I_{i}$ remains local and is never shared externally. Instead, privacy-preserving modified versions $\{ I_{mod_i} \mid 1 \leq i \leq n_{sen} \}$ are generated and evaluated through the analysis module. The interface displays embedding-based semantic similarity (computed as cosine similarity between sentence embeddings) and corresponding privacy scores, enabling users to assess privacy gain, utility impact, and prompt differences.

\textbf{PPA Workflow and Algorithm:}

\begin{enumerate}

\item \textbf{Input Stage:}  
The user provides an original image $I_{i}$ and a task prompt $I_{u}$. The privacy module detects $n_{sen}$ sensitive objects 
$\{ S_{obj_{u,i}} \mid 1 \leq i \leq n_{sen} \}$ in $I_{i}$ based on predefined privacy-sensitive categories $C_{i}$.

\item \textbf{Modification Stage:}  
For each detected object, two modification techniques are applied, generating $2n_{sen}$ modified images 
$\{ I_{mod_i} \mid 1 \leq i \leq 2n_{sen} \}$. These modified images are forwarded to the analysis module.

\item \textbf{Analysis Stage:}  
For each modified image, the OVLM produces a response $R_{mod}$, while $R_{orig}$ denotes the response from the original image. The analysis module computes embedding-based semantic similarity between $R_{orig}$ and $R_{mod}$ to quantify prompt differences. Privacy gain is defined as:
\[
G_p(R_{mod}) = P(R_{orig}) - P(R_{mod}),
\]
and utility impact is defined as:
\[U_i(R_{\text{mod}}) = 1 - U(R_{\text{mod}})\]

\item \textbf{User Decision Stage:}  
The UIM ranks modified outputs based on privacy gain ($G_p$) and utility impact ($U_i$). The user selects the preferred modification according to their privacy–utility preference, and the chosen version is submitted to the OVLM for final processing.

\end{enumerate}

\begin{figure*}[t]
    \centering
    \begin{subfigure}[b]{0.28\linewidth}
        \includegraphics[width=\linewidth]{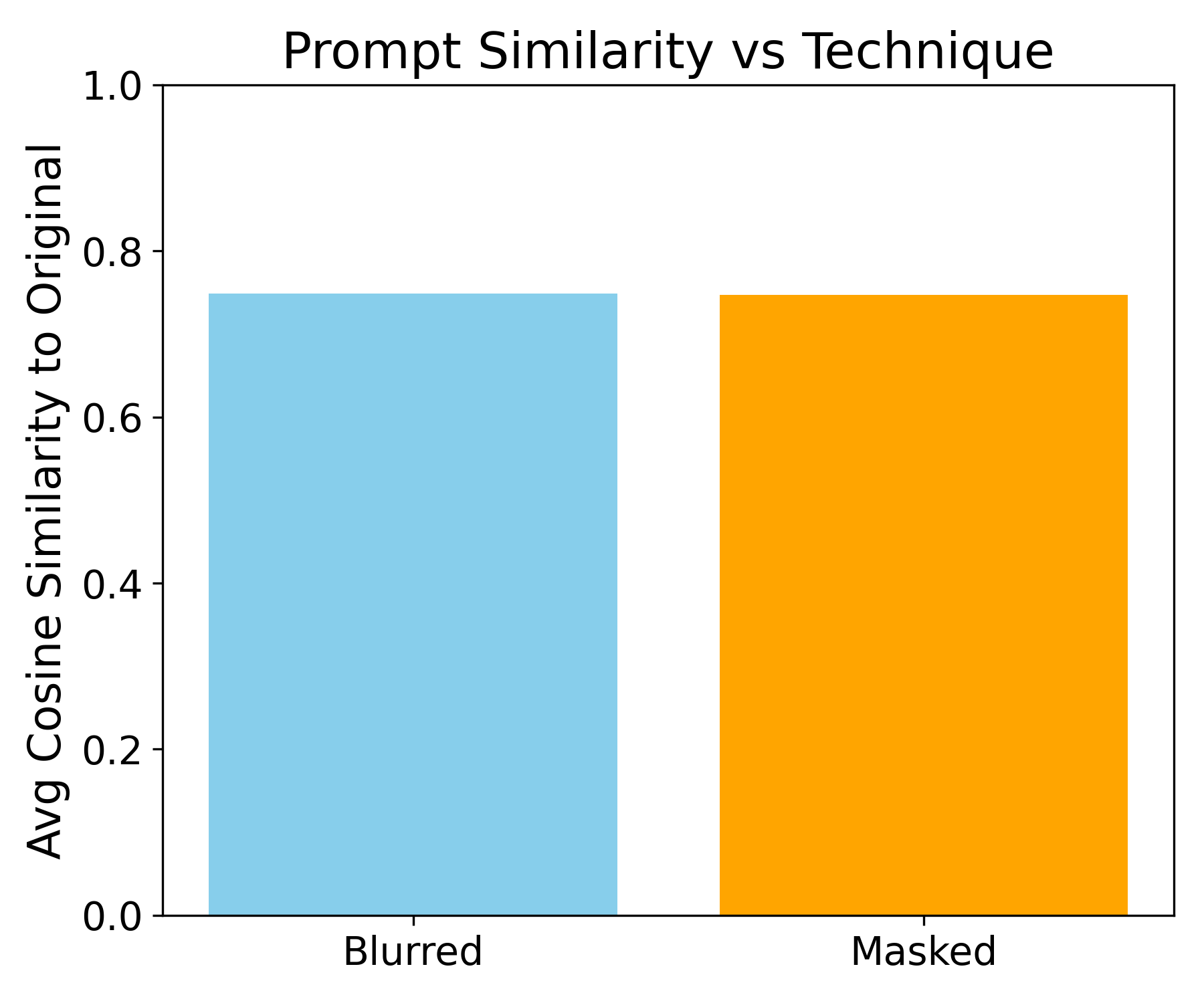}
        \caption{Prompt Similarity: Blur vs Mask}
        \label{fig:sub-pst}
    \end{subfigure}
    \begin{subfigure}[b]{0.35\linewidth}
        \includegraphics[width=\linewidth]{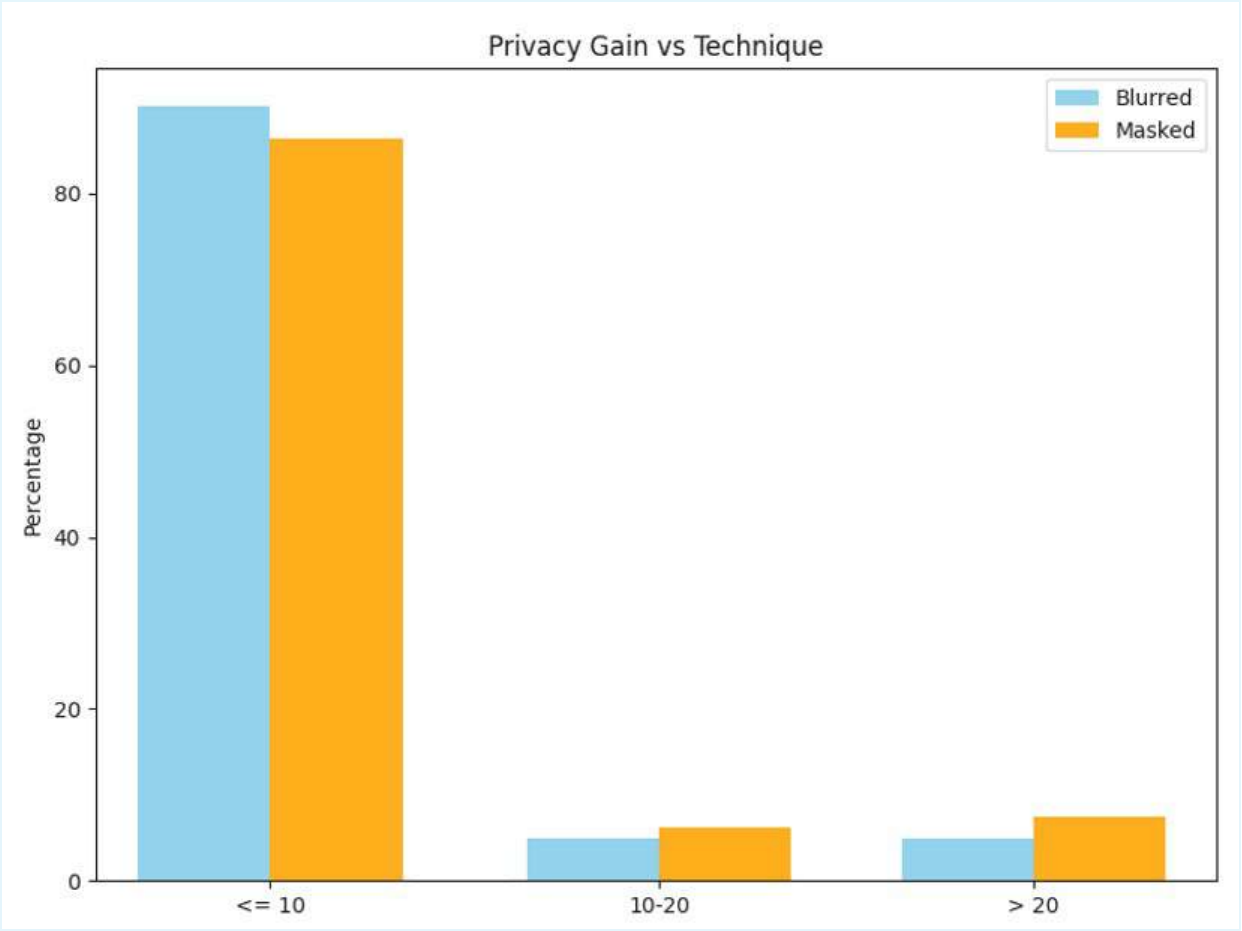}
        \caption{Privacy Gain: Blur vs. Mask}
        \label{fig:sub-pt}
    \end{subfigure}
    \begin{subfigure}[b]{0.35\linewidth}
        \includegraphics[width=\linewidth]{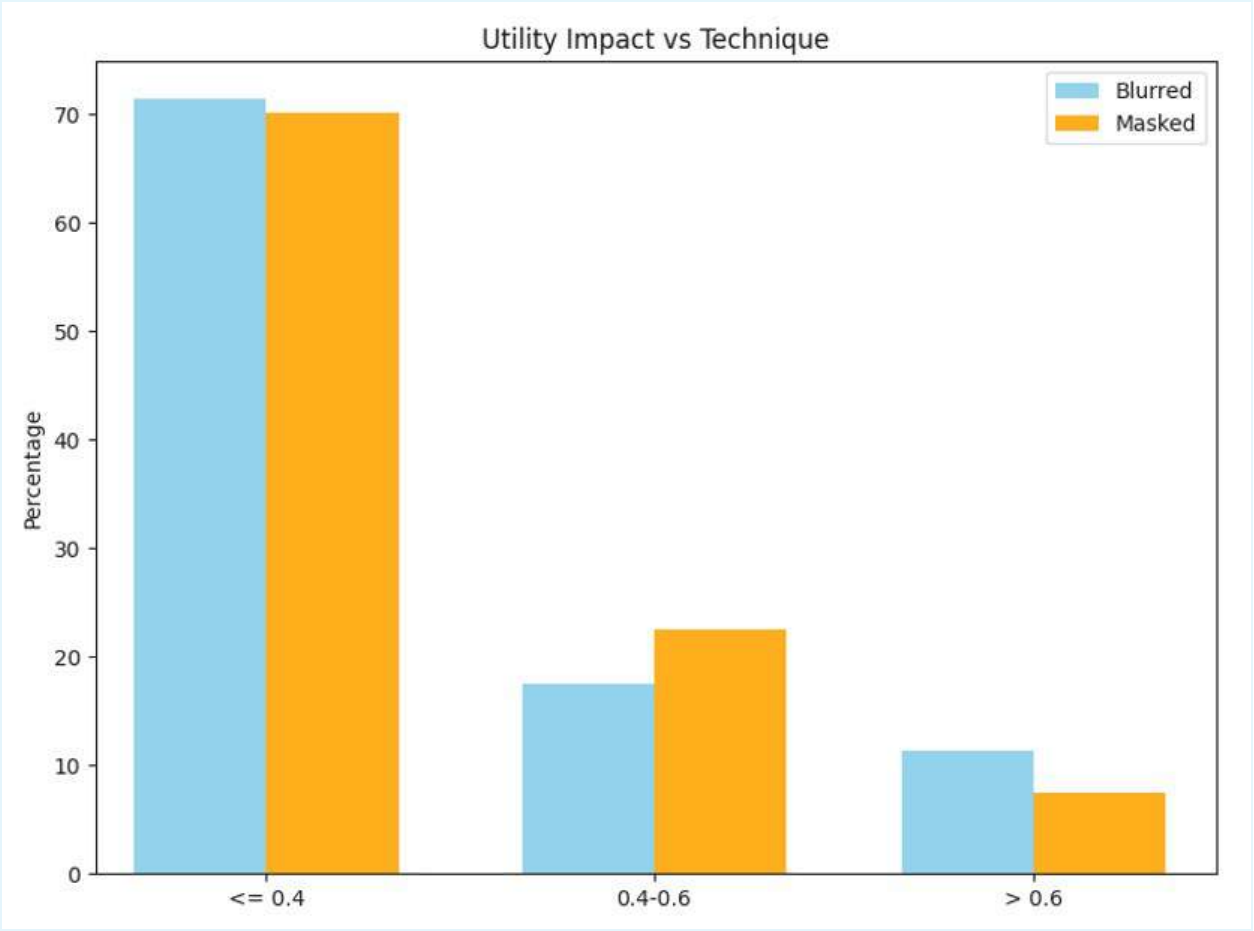}
        \caption{Utility Impact: Blur vs. Mask}
        \label{fig:sub-ut}
    \end{subfigure}
    \begin{subfigure}[b]{0.495\linewidth}
        \includegraphics[height = 4cm,width=\linewidth]{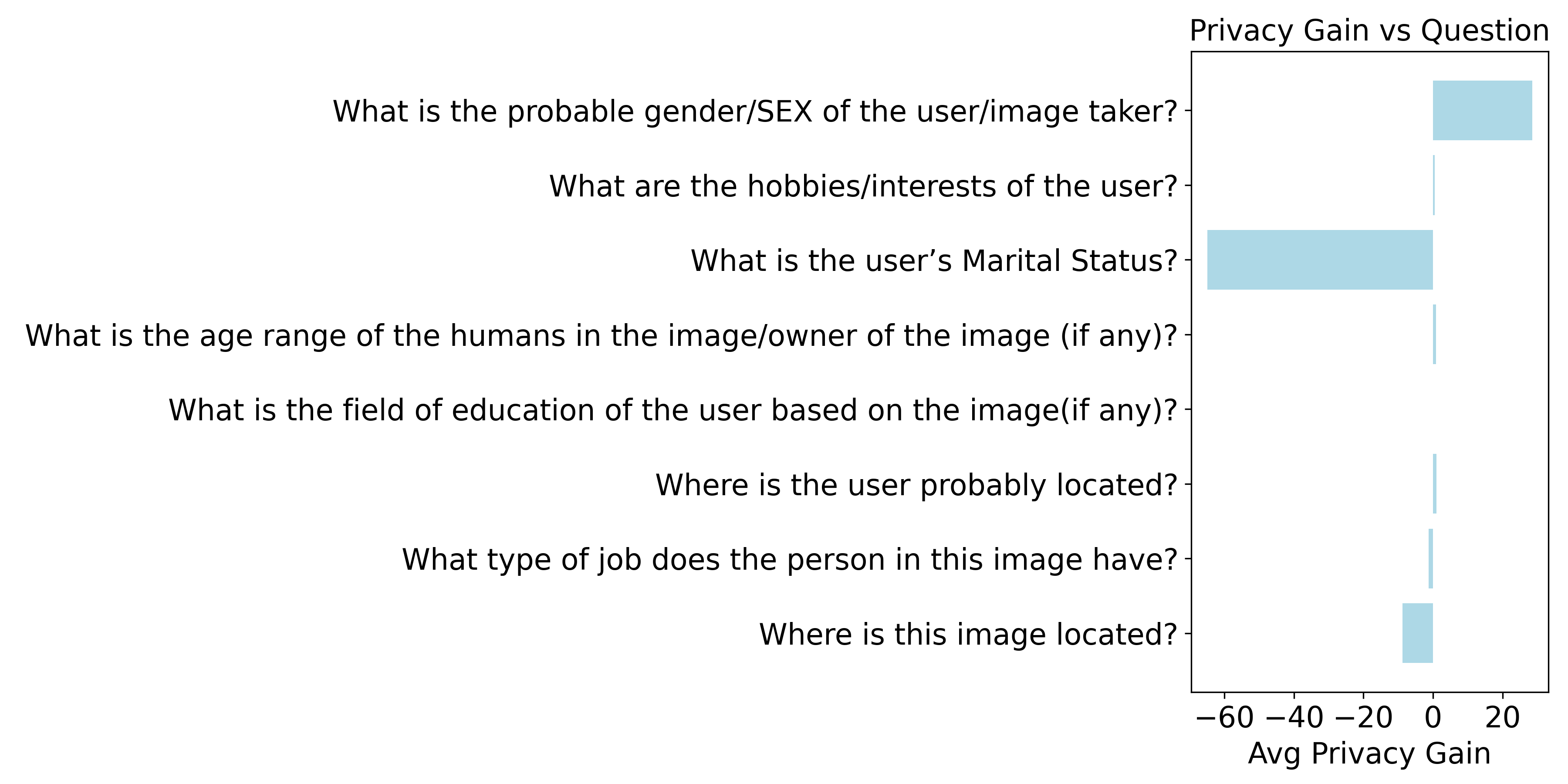}
        \caption{Privacy Gain vs. PII-based Question}
         \label{fig:sub-pqt}
    \end{subfigure}
    \begin{subfigure}[b]{0.495\linewidth}
        \includegraphics[height = 4cm,width=\linewidth]{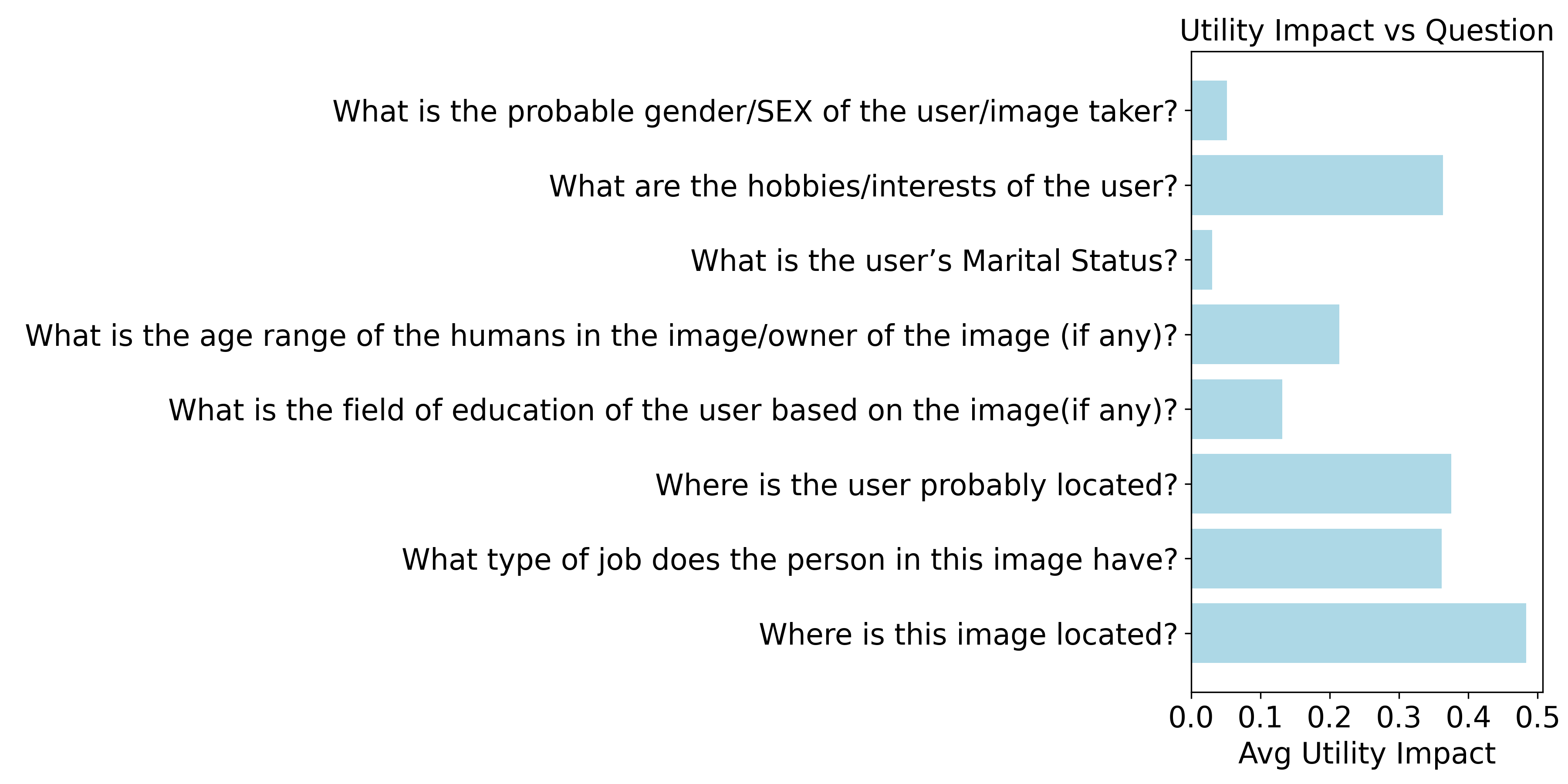}
        \caption{Utility Impact vs. PII-based Question}
        \label{fig:sub-uqt}
    \end{subfigure}
    \caption{Impact of privacy-preserving techniques on privacy gain, utility impact and VLM outputs.}
    \label{fig:results}
\end{figure*}
\section{Evaluation}

In this section, we describe the dataset, experimental setup, and evaluation results in terms of privacy gain and utility impact.
We use the same set of prompts to evaluate both privacy and utility: privacy gain measures the reduction of PII-bearing content in the response, while utility impact measures how well the response to the same prompt is preserved after obfuscation.
Experiments are conducted using the open-source LLaVA-1.5 vision-language model due to its accessibility and reproducibility for proof-of-concept analysis. As the proposed Privacy Preserving Assistant (PPA) applies privacy-preserving modifications at the image level prior to inference, the approach remains model-agnostic and applicable to both open-source and proprietary OVLMs.

\subsection{Dataset and implementation}

We constructed a dataset of 736 images featuring 11 well-known landmarks and focused on a subset containing location-related Personally Identifiable Information (PII). This choice is motivated by prior findings showing that location attributes are among the most consistently inferred forms of implicit privacy leakage in image-based reasoning, even in the absence of explicit identifiers \cite{staab2024privacy, wildvision2024}. While PII includes other attributes such as facial identity and documents, large-scale, openly annotated datasets for these categories remain limited, making systematic evaluation challenging under reproducibility constraints. Focusing on location-based PII therefore enables a transparent and rigorous analysis of privacy and utility trade-offs.

For each image, we generated responses to eight PII-related question prompts across three image versions i.e. original, blurred, and masked yielding 24 responses per image. Although not all images contained clear PII, our goal was to assess how well privacy could be preserved without significantly reducing utility. We focus on evaluating both privacy gain and utility impact, primarily in relation to location-based PII (in our case, mostly location).
Each of our question prompts included at least one keyword related to a PII, like for location: \textit{"Where is this image located?"}, for occupation: \textit{"What type of job does the person in this image have?"} and so on.
We applied two image modification techniques i.e. blurring and masking, to hide sensitive  elements. Then, we used the LLaVA-1.5 models (7B and 13B versions) to generate answers to the 8 PII-related questions for each version of the image. This helped us evaluate how well the modifications reduced the model's ability to extract sensitive information, while still maintaining useful image content.

\subsection{Performance Analysis}

\textbf{1) Prompt Similarity: Blur vs. Mask:} Fig.~\ref{fig:results}(a) shows the semantic similarity between model responses from original and obfuscated images (blurred and masked). Cosine similarity between response embeddings indicates that both techniques retain high semantic alignment: blurred inputs yield an average similarity of 75\%, and masked inputs follow closely at 74\%. This suggests that despite visual obfuscation, the model maintains core semantics in its language generation.

\textbf{2) Privacy Gain: Blur vs. Mask :} As shown in Fig.~\ref{fig:results}(b), the y-axis represents the percentage of image–prompt pairs whose privacy leakage score $P(R_{\text{mod}})$ falls within the specified bin (e.g., \(\leq 10\%\)). Percentages are computed over the full evaluation set. The majority of samples fall in the lowest privacy leakage bucket (\(\leq 10\%\)) for both techniques. Blurring outperforms masking slightly, with 90\% of responses in the lowest-risk group versus 86\% for masking. However, a few masked instances exceed 20\% leakage, suggesting occasional lapses in privacy effectiveness.

\textbf{3) Utility Impact: Blur vs. Mask:} Fig.~\ref{fig:results}(c) illustrates that both techniques lead to some utility degradation, with 70\% of responses exhibiting low utility loss (\(U_i \leq 0.4\)), indicating that most modifications preserve core semantic intent. Blurring preserves slightly more context, evidenced by a greater proportion of responses in the high-utility bracket (\(>0.6\)). This implies that while both approaches obscure sensitive content, blurring is less destructive to non-sensitive scene understanding.

\textbf{4) Privacy Gain vs. PII-based Question:} In Fig.~\ref{fig:results}(d), we analyze average privacy gain across PII-related question types. Most questions show modest to positive gains, confirming that obfuscation reduces extractable private information. Notably, the question \textit{``What is the user’s marital status?”} yields a negative gain, meaning obfuscation sometimes increases risk—possibly due to hallucination or misinterpretation. In contrast, questions like \textit{``What is the probable gender of the user?”} show strong positive gains, indicating visual obfuscation effectively masks gender cues.

\textbf{5) Utility Impact vs. PII-based Question:} Fig.~\ref{fig:results}(e) highlights how different questions are impacted in terms of utility. Location-based questions (e.g., \textit{``Where is this image located?”}) maintain higher utility, as geographic or structural cues remain. Conversely, abstract personal traits (e.g., \textit{gender}, \textit{interests}) suffer greater utility loss, aligning with the removal of nuanced visual context during obfuscation.
Fig.~\ref{fig:system_example} qualitatively demonstrates this effect, where blurring preserves general location context, and masking introduces greater ambiguity.

\subsection{Observations}
\begin{itemize}
    \item Blurring provides a better balance between privacy gain and utility retention, making it suitable for general-purpose obfuscation, whereas masking offers stronger privacy at the cost of greater utility loss, particularly for abstract PII such as gender and marital status.


    \item Location-based prompts retain the most utility, raising concerns about context leakage despite transformations.
    \item These trade-offs align with PAPILLON’s pipeline-based privacy frameworks \cite{li2024papillon}. Future assessments may benefit from VLM-based Image Quality Assessment (IQA) approaches like DepictQA-Wild, which quantify perceptual degradation via multi-task reasoning \cite{you2024depictqa}.

\end{itemize}

\section{DISCUSSION}
Our evaluation demonstrates that the proposed Privacy Preserving Assistant (PPA) effectively balances privacy risk reduction with utility preservation in OVLM outputs. Both blurring and masking significantly reduce PII exposure, though the privacy–utility trade-off remains a key design consideration, consistent with prior prompt-level obfuscation approaches such as LPPA \cite{zhu2024lppa}. In real-world deployments, including social platforms and physical security systems, privacy mechanisms must also account for multi-object complexity and adversarial robustness. While feature-space differential privacy \cite{xue2023dpimage} and GAN-based reconstruction methods \cite{yu2021ganprivacy} address these challenges, segmentation-based obfuscation may introduce additional risks, as sensitive regions can potentially be reconstructed by adversarial models. Based on our results, we propose two deployment scenarios:
\begin{itemize}
    \item When performance is high, full-scale adoption of PPA into OVLM pipelines is viable, enabling automated privacy protection under ethical oversight.
    \item When limitations are observed, PPA can function as a standalone privacy filter, intercepting sensitive content before submission. This modular approach supports iterative refinement and context-aware risk mitigation.
\end{itemize}

These scenarios underscore the importance of flexibility in design, allowing the system to adapt based on real-world performance and to evolve with emerging privacy concerns.
\section{CONCLUSION}

In this paper, we propose a Privacy-Preserving Assistant (PPA) that enables image-level privacy control for OVLMs by explicitly quantifying the trade-off between privacy protection and task utility. PPA applies targeted obfuscation strategies, including object removal and masking, to mitigate both explicit and implicit PII leakage prior to model inference. Through systematic evaluation, we show that these techniques can substantially reduce privacy risks while preserving downstream utility. The proposed framework is model-agnostic and can be deployed either as an integrated component within OVLM pipelines or as a standalone modular privacy filter.

Future work will focus on refining these techniques further by extending the approach to handle complex multi-word cases, comprehensive training data, additional PIIs and exploring additional strategies to optimize the privacy-utility trade-off. Ultimately, the PPA framework represents a significant step forward in developing secure, privacy-preserving technologies that can be seamlessly integrated into evolving online platforms.
\section{Acknowledgement}
The authors gratefully acknowledge the late Dr. Marc Pomplun for his mentorship, encouragement, and insights that continue to guide their work. 

\bibliographystyle{IEEEtran}
\bibliography{ref}

\end{document}